\documentclass[11pt]{article}

\usepackage[preprint]{acl}

\usepackage{times}
\usepackage{latexsym}
\usepackage{mdframed,lipsum}
\usepackage[table]{xcolor}

\usepackage[T1]{fontenc}
\usepackage[utf8]{inputenc}

\usepackage{microtype}

\usepackage{inconsolata}

\usepackage{graphicx}

\usepackage{graphicx}
\usepackage{amsmath}
\usepackage{amssymb}
\usepackage{booktabs}
\usepackage{multirow}
\usepackage{array}
\usepackage{enumitem}
\usepackage{url}
\usepackage{xcolor}
\usepackage{hyperref}
\hypersetup{hidelinks}
\usepackage{tikz}
\usepackage{svg}
\usetikzlibrary{patterns}
\usepackage{pgfplots}
\usepackage{pgfplotstable}
\pgfplotsset{compat=1.18}

\usepackage{pgfplots}
\pgfplotsset{compat=1.18}
\definecolor{ourblue}{RGB}{55, 126, 184}
\definecolor{bline1}{RGB}{200, 200, 200}
\definecolor{bline2}{RGB}{160, 160, 160}
\definecolor{bline3}{RGB}{120, 120, 120}
\definecolor{bline4}{RGB}{80, 80, 80}
\usepackage{soul}
\usepackage[utf8]{inputenc}
\usepackage{graphicx}
\usepackage{xcolor}
\usepackage{tabularx}
\definecolor{highlightgreen}{RGB}{144, 238, 144}
\definecolor{highlightpink}{RGB}{255, 182, 193}
\definecolor{highlightorange}{RGB}{255, 200, 100}
\newcommand{\hlg}[1]{\sethlcolor{highlightgreen}\hl{#1}}
\newcommand{\hlp}[1]{\sethlcolor{highlightpink}\hl{#1}}
\newcommand{\hlo}[1]{\sethlcolor{highlightorange}\hl{#1}}
\definecolor{highlightgreen}{RGB}{144, 238, 144}
\definecolor{highlightpink}{RGB}{255, 182, 193}
\definecolor{highlightorange}{RGB}{255, 200, 100}
\definecolor{highlightblue}{RGB}{173, 216, 230}
\definecolor{highlightyellow}{RGB}{255, 255, 170}
\definecolor{highlightpurple}{RGB}{216, 191, 216}
\newcommand{\hlk}[1]{\sethlcolor{highlightblue}\hl{#1}}
\newcommand{\hly}[1]{\sethlcolor{highlightyellow}\hl{#1}}
\newcommand{\hlm}[1]{\sethlcolor{highlightpurple}\hl{#1}}

\newcommand{\Lcal}{\mathcal{L}}

\newcommand{\nostd}[1]{#1} % for CT2Rep (std unavailable)

\newcommand{\sd}[2]{%
  #1\ensuremath{_{\textcolor{black!55}{\scriptscriptstyle\!\pm\,#2}}}%
}

\newcommand{\bsd}[2]{%
  \textbf{#1}\ensuremath{_{\pmb{\scriptscriptstyle}\!\pm\,#2}}%
}

\title{Generating Reports or Repeating Templates? Measuring and \\ Mitigating Template Collapse in 3D CT Report Generation}

\author{
  \textbf{Tom Maye-Lasserre\textsuperscript{1}\thanks{Equal contribution.}},
  \textbf{Yitong Li\textsuperscript{1,2,3}\footnotemark[1]},
  \textbf{Bailiang Jian\textsuperscript{1,2,3}\footnotemark[1]},
  \\
  \textbf{Morteza Ghahremani\textsuperscript{1,2,3}},
  \textbf{Benedikt Wiestler\textsuperscript{1,2,3}},
  \textbf{Christian Wachinger\textsuperscript{1,2,3}} \\ 
  \textsuperscript{1}Technical University of Munich (TUM), Germany \\
  \textsuperscript{2}TUM Hospital, Germany \\
  \textsuperscript{3}Munich Center for Machine Learning (MCML), Germany
  \\
 \small{
   \textbf{Correspondence:} \href{mailto:yi_tong.li@tum.de}{yitong\_li@tum.de}
 }
}
\begin{document}
\maketitle
\begin{abstract}
Modern 3D medical vision-language models (VLMs) can generate fluent radiology-style text while 
exhibit critically low pathology detection and output diversity, collapsing to generic templates that under-report rare yet critical findings. 
% repeatedly emitting generic report patterns that omit patient-specific abnormalities.
We identify this failure mode as \emph{Template Collapse}.
% : high surface overlap with reference reports coexists with low pathology detection, reduced output diversity, and strong compression of critical findings.
This failure stems from the unique constraints of 3D medical imaging, e.g., limited data, severe label imbalance, and weak signals from volumetric encoders. 
Under these constraints, text-generation objectives encourage shortcut learning and fluent but weakly grounded reports.
We systematically diagnose the Template Collapse through clinical fidelity, output diversity, normal-template bias, and rare-finding survival.
To mitigate it, we propose \emph{CLarGen}, a decoupled framework that separates \emph{what to say} (clinical detection) from \emph{how to say it} (language synthesis).
CLarGen uses (i) a Latent Query Transformer for multi-label pathology detection, (ii) pathology-guided retrieval for clinically matched exemplars, and (iii) a medical language model to synthesize the final report from detected findings and retrieved context.
Across state-of-the-art 3D CT report generation baselines, CLarGen mitigates Template Collapse and substantially improves clinical accuracy (macro-F1 0.487 vs. 0.189; CRG 0.472 vs. 0.368) while maintaining fluent reporting. 
Our results suggest that explicit, measurable clinical grounding is essential for template-collapse-resistant 3D CT report generation. Code is available at \url{https://github.com/ai-med/CLarGen}.

\end{abstract}

\begin{figure*}[t]
    \centering
    \includegraphics[width=\linewidth]{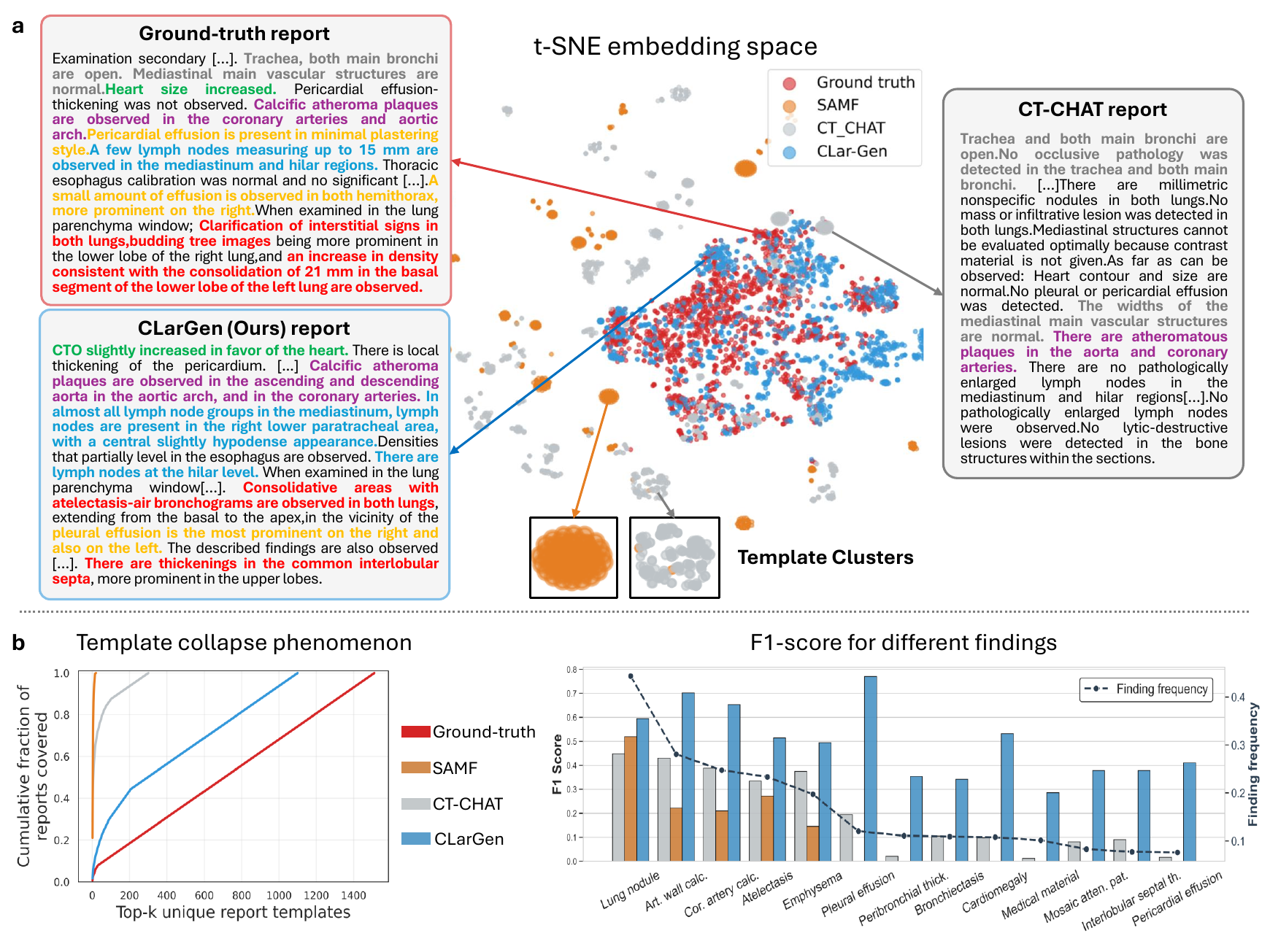}
    \vspace{-0.8em}
    % \caption{(a) Color-coded text shows that CLarGen reports align with ground-truth pathologies, whereas baseline VLMs produce generic descriptions with limited findings. This behavior is reflected in the t-SNE space: CLarGen outputs are broadly distributed and closely match the ground-truth manifold, while other methods form tight clusters corresponding to a few dominant templates. (b) Left: the cumulative coverage of reports as a function of the number of unique templates further illustrates this template collapse phenomenon in standard VLMs, generating reports from a limited set of generic patterns. Right: this leads to clinical bias: common findings are over-reported while rare but critical pathologies are frequently missed. In contrast, CLarGen reduces template dependence and maintains detection performance across findings of varying prevalence.}

    %Bailiang's version
    \vspace{-0.5em}
    \caption{\textbf{Template Collapse in 3D CT report generation.} (a) Generated reports from VLM baselines form compact clusters in report-embedding space, indicating that many scans are mapped to a small set of semantically similar templates rather than patient-specific descriptions. The color-coded text shows that CLarGen preserves pathology-specific findings aligned with the ground-truth report, while baselines produce generic descriptions with limited findings. (b) Left: cumulative report coverage as a function of the number of unique templates further confirms that a few repeated report patterns (Appendix~\ref{app:dominant_templates}) account for a large fraction of baseline outputs. Right: this produces a clinical bias: common findings are over-reported, while rare but critical pathologies are frequently missed. CLarGen mitigates these effects, producing a broader report distribution and maintaining detection performance across findings of varying prevalence.}
    \label{fig:hook_templates_rarity}
    \vspace{-0.5em}
\end{figure*}

%%%%%%%%%%%%%%

% \begin{figure}[t]
%     \centering
%     \includegraphics[width=\linewidth,height=4.1cm,keepaspectratio]{F1_score_per_pathology.pdf}
%     \caption{\textbf{Overcoming template collapse for robust clinical detection.} Standard VLMs default to frequent normal findings, leading to catastrophic F1 drops on rare but critical anomalies. CLarGen explicitly decouples detection from generation, maintaining robust accuracy across all pathology frequencies.}
%     \label{fig:rarity_f1}

% \end{figure}
%%%%%%%%%%%%%%%%%%%

\section{Introduction}
\label{sec:intro}

Automated radiology report generation for 3D Computed Tomography (CT) is a high-stakes task, where the precise identification and description of pathological findings is the primary goal. 
A linguistically fluent report that omits a lung nodule or hallucinates a non-existent effusion is not merely suboptimal but potentially harmful. Consequently, the accuracy of \emph{what} is reported must take precedence over the elegance of \emph{how} it is written.
% However, current approaches, which predominantly fine-tune decoder-only vision-language models (VLMs) via next-token prediction~\cite{hamamci2024ct2rep,hamamci2026generalist,hosseini2025samf,jiang2025hulu}, frequently fail this clinical priority. 

However, our empirical study of recent vision-language models (VLMs) for 3D CT report generation~\cite{hamamci2024ct2rep,chen2025largelanguagemodelregionguided,hamamci2026generalist,hosseini2025samf,jiang2025hulu} showed a striking failure mode: instead of producing patient-specific reports, they tend to converge to a small set of generic report templates. We term this phenomenon \emph{Template Collapse}.
Fine-tuned with standard next-token prediction, these models can achieve strong conventional NLP fluency metrics (BLEU, ROUGE) while performing poorly on clinical detection (F1-score) and exhibiting severely limited report diversity. 
As illustrated in Fig.~\ref{fig:hook_templates_rarity}, their generated reports are linguistically plausible yet diagnostically uninformative, often clustering around a few dominant templates.
This template-driven behavior biases generation toward common findings, reduces report diversity, and frequently suppresses rare yet clinically significant abnormalities that are critical for diagnosis and downstream medical decision-making.
% and failing to capture the rare but clinically critical abnormalities that drive medical decision-making.

To our knowledge, this failure mode has not been explicitly characterized in prior work. While related issues may already affect report generation in the more widely studied 2D setting~\cite{bannur2024maira2groundedradiologyreport,li2025reevalmed}, they are likely amplified in 3D CT.
Compared to 2D chest radiography, available 3D CT report datasets are substantially smaller~\cite{hamamci2026generalist,hwang2025ai}, labels are extremely imbalanced with most findings absent in any given scan, and 3D visual encoders must align large volumes of imaging information with fine-grained clinical semantics~\cite{blankemeier2026merlin}. Subtle abnormalities are embedded within millions of voxels, yet the model must simultaneously identify findings, reason about their relationships, and generate a coherent report. Together, these factors create a low-signal, high-imbalance learning regime where standard text-optimization objectives provide weak incentives for diagnostic accuracy while strongly rewarding the reproduction of generic normal-report templates.
To make this failure mode measurable, we evaluate 3D CT report generation beyond surface-level text overlap. We treat Template Collapse as a joint degradation in clinical fidelity, report diversity, semantic spread, normal-bias similarity, dominant-template concentration, and rare-finding survival.
This diagnostic view exposes failures that ROUGE/BLEU can obscure: a model may match common report phrasing while repeatedly omitting patient-specific abnormalities.
Motivated by this diagnosis, we propose CLarGen, a decoupled framework that separates clinical detection from linguistic synthesis for 3D CT report generation.
Recent approaches largely rely on end-to-end LLM fine-tuning, where pathology recognition and report generation are jointly optimized within a single model~\cite{di2025ct,chen2025dia,deng2025mvketr}.
In contrast, CLarGen explicitly disentangles into three stages: clinical perception, knowledge retrieval, and report synthesis. 
% First, a Latent Query Transformer for multi-label pathology classification explicitly models inter-finding correlations; 
% Second, the predicted pathology sets guide the retrieval of clinically consistent exemplar reports from a reference database. Finally, a medical LLM integrates the detected findings and retrieved context into a coherent radiology report. 
% Bailiang's version
Our main contributions are threefold:
(1) We identify and characterize \emph{Template Collapse}, a failure mode in 3D CT report generation where VLMs produce fluent but repetitive reports that suppress patient-specific findings.
(2) We systematically measure this failure through complementary diagnostic aspects.
% of clinical fidelity, lexical diversity, normal-template bias, dominant-template concentration, and rare-finding survival.
(3) We propose CLarGen, a pathology-guided retrieval-augmented framework that mitigates Template Collapse 
% by separating clinical evidence detection from language synthesis, 
and substantially improves pathology detection while preserving fluent and clinically realistic reporting.

\section{Related Work}
\label{sec:related}

\paragraph{Radiology report generation.}
Radiology report generation (RRG) has been studied primarily on 2D chest X-rays, where large-scale paired datasets and mature 2D visual encoders have enabled strong vision-language models~\cite{bannur2024maira2groundedradiologyreport,wu2023radfm,sun2025fact,liu2025enhanced,parikh2026measuring,liu2026priorrg}.
% such as MAIRA-2~\cite{bannur2024maira2groundedradiologyreport} and RadFM~\cite{wu2023radfm}, FactMM-RAG~\cite{sun2025fact}. 
Recently, the release of CT-RATE~\cite{hamamci2026generalist}, the first large-scale paired 3D CT-report dataset, has accelerated research in 3D RRG. However, 3D RRG remains substantially more challenging due to limited paired data, the extreme sparsity of pathological findings within high-dimensional volumes~\cite{hwang2025ai}, and weaker semantic alignment between volumetric encoders and clinical language compared with 2D models.

\paragraph{Vision-language models for 3D CT.}
The prevailing 3D RRG paradigm couples a 3D image encoder
% ---CT-CLIP~\cite{hamamci2026generalist}, HLIP~\cite{zhao2025towards}, Curia~\cite{dancette2025curiamultimodalfoundationmodel}, or 
with a decoder-only LLM trained end-to-end via autoregressive next-token prediction, including CT2Rep~\cite{hamamci2024ct2rep}, CT-CHAT~\cite{hamamci2026generalist}, M3D~\cite{bai2024m3d}, and 3D-CT-GPT~\cite{chen2024ctgpt} adapting the LLaVA recipe~\cite{liu2023visual} to volumetric inputs. 
Subsequent work has explored multi-scale 2D/3D fusion~\cite{hosseini2025samf}, region- and anatomy-guided grounding~\cite{chen2025largelanguagemodelregionguided,kyung2025medregionct}, interactive segmentation~\cite{gu2025interactive}, generalist medical VLMs~\cite{jiang2025hulu,blankemeier2026merlin}, and knowledge-based report augmentation~\cite{baharoon2025exploring}.
Recent work incorporates explicit clinical signals into report generation: CT-AGRG~\cite{di2025ct} conditions generation on a learned abnormality vector, Dia-LLaMA~\cite{chen2025dia} introduces disease-aware prompting, and MvKeTR~\cite{deng2025mvketr} proposes multi-view feature alignment. 
However, these methods are predominantly optimized with autoregressive language modeling objectives. We observe that, in the low-data and highly imbalanced 3D setting, such objectives can bias models toward frequent ``normal'' report templates rather than subtle pathological findings, leading to a failure mode we term \emph{template collapse}.
% On 2D chest X-ray, LaB-RAG~\cite{song2024labrag} shows that classification labels can boost RAG with general-purpose LLMs, though its setting differs from ours in modality, signal density, and encoder maturity.
% General retrieval-augmented generation (RAG)~\cite{lewis2020rag} grounds LLMs in factual context for knowledge-intensive tasks, but has not yet been adapted to 3D CT reporting with clinical-label guidance.
% CLarGen differs along three axes: \textbf{(i)} detection is fully decoupled from synthesis via an explicit pathology classifier optimized with a clinical objective; \textbf{(ii)} retrieval is \emph{pathology-guided}, ranking exemplars by predicted-label consistency rather than embedding similarity alone---essential when 3D encoders remain weakly aligned with clinical semantics; and \textbf{(iii)} the medical LLM is kept \emph{frozen}, preserving its pretrained clinical reasoning and avoiding the dataset-specific shortcuts that drive template collapse.
CLarGen differs in three key aspects: \textbf{(i)} perception and synthesis are explicitly decoupled through a pathology classifier trained with clinical supervision; \textbf{(ii)} retrieval is pathology-guided, ranking exemplars by predicted label consistency rather than embedding similarity alone; and \textbf{(iii)} the medical LLM remains frozen, preserving pretrained clinical reasoning while reducing overfitting to frequent report templates.

\section{Method}
\label{sec:method}

\begin{figure*}[t]
    \centering
    \includegraphics[width=0.95\linewidth]{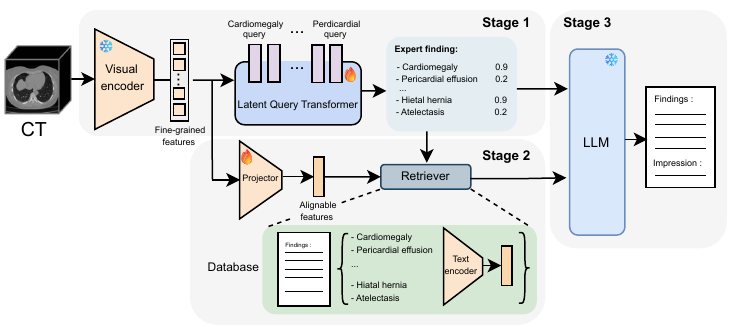}
    \caption{CLarGen is a three-stage medical-grounded pipeline for CT report generation.}
    \label{fig:schema}
    \vspace{-0.5em}
\end{figure*}

% \subsection{CLarGen: Expert-Guided Retrieval-Augmented Generation}
% \label{ssec:clargen} 

% Bailiang's version
We now introduce CLarGen for 3D CT report generation.
As motivated in the introduction, Template Collapse refers to the tendency of autoregressive VLMs to produce fluent but repetitive reports that omit patient-specific findings; we empirically analyze this failure mode in Sec.~\ref{sec:results}.
CLarGen is designed as a mitigation strategy: instead of asking one model to both detect subtle 3D findings and verbalize them, it makes clinical detection explicit before language synthesis.
As shown in Fig.~\ref{fig:schema}, CLarGen decomposes report generation into three stages: (1) explicit pathology modeling and classification from 3D features, (2) label-aware retrieval of exemplar reports, and (3) report synthesis with a medical language model.
This medical-grounded pipeline first converts 3D CT volumes into fine-grained visual tokens, predicts calibrated pathology scores to anchor clinically relevant retrieval, and finally conditions the LLM to consolidate this information into a final report.

% Original Version
% Radiology report generation from 3D CT necessitates multiple sub-tasks, including pathology detection, modeling of inter-finding dependencies, and structured text synthesis. As shown in Fig.~\ref{fig:schema}, CLarGen addresses this by decomposing the problem into three sequential stages: (1) explicit pathology modeling and classification from 3D features, (2) label-aware retrieval of exemplar reports, and (3) report synthesis with a medical language model.
% This medical-grounded pipeline first converts 3D CT volumes into fine-grained visual tokens, predicts a set of calibrated pathology scores to anchor the retrieval of relevant medical context, and finally conditions the LLM to consolidate this information into a final report.

\paragraph{Stage 1: Latent query transformer for clinical perception.}
Given a frozen 3D image encoder output $\mathbf{Z} \in \mathbb{R}^{N \times D}$, global pooling risks obscuring small abnormalities such as small nodules. We therefore introduce a Latent Query Transformer (LQT) with $C=18$ learnable pathology queries $\mathbf{Q} \in \mathbb{R}^{C \times D}$. Each query represents a specific pathology (e.g., pleural effusion, atelectasis) and attends to $\mathbf{Z}$ via cross-attention to extract pathology-specific evidence, while self-attention among queries models clinical correlations between findings. The refined queries are projected to multi-label logits.
To address extreme class imbalance, we optimize a focal loss~\cite{lin2017focal}:
% Instead of performing global pooling on the long frozen sequence $\mathbf{Z} \in \mathbb{R}^{1492 \times 768}$, which risks diluting focal abnormalities (e.g., small nodules), we design a query transformer head. We initialize $C=18$ learnable latent queries $\mathbf{Q} \in \mathbb{R}^{C \times D}$, each representing a specific pathology. These queries flow through a Transformer Decoder, utilizing (i) \emph{cross-attention} to extract pathology-specific visual evidence from $\mathbf{Z}$, and (ii) \emph{self-attention} to explicitly model clinical correlations between findings (e.g., pleural effusion and atelectasis). The refined queries are projected to logits and trained using Focal Loss~\cite{focalloss} : Focal loss because pathologies extreme class imbalance as most patient are normal, so this loss handles abnormal pathologies to be considered. 
\begin{equation}
\Lcal_{\text{FL}} = -\sum_{c=1}^{C} \alpha_c\,(1-p_{t,c})^{\gamma}\,\log(p_{t,c}),
\end{equation}
where $\alpha_c$ is the inverse class frequency and $p_{t,c}$ the predicted probability for class $c$. 
This stage provides direct supervision on clinically meaningful targets, decoupled from language modeling.
% This supervised training allows us to have direct supervision and with "medical" optimization using the "medical aware" focal loss over main pathologies instead of "undirect" supervised training with VLMs that uses text optimization over the next token. Not forcing the model to care about medical but text instead. This bias we introduce provides way better results when limited/ unbalanced data. 

\paragraph{Stage 2: Pathology-Guided Retrieval.}
We retrieve $k$ exemplar reports that are visually similar to the query CT, clinically consistent with the predicted pathologies from Stage~1, and collectively cover all high-confidence findings.
Let $\mathcal{B}=\{(r_i,\mathbf{y}_i,\mathbf{t}_i)\}_{i=1}^{M}$ denote a report collection, where
$r_i$ is the report text,
$\mathbf{y}_i\in\{0,1\}^C$ is the multi-label pathology vector for $C$ different labels, and
$\mathbf{t}_i\in\mathbb{R}^d$ a precomputed text embedding.
Given a query CT, we compute an image-to-text retrieval embedding $\mathbf{v}\in\mathbb{R}^d$
using a projector trained contrastively with CT and text pairs, and pathology probabilities $\mathbf{p}\in[0,1]^C$ from the LQT.
% \begin{enumerate}
% \item \textbf{Search space reduction.}
% We first reduce the search space to a candidate set $\mathcal{P}$ of the top $N=2,000$ reports based on the cosine similarity between the query embedding $\mathbf{v}$ and text embeddings $\mathbf{t}_i$.
% \item \textbf{Fusing visual retrieval and pathology alignment.}
To ensure the retrieved context reflects the detected pathologies, we compute a clinical alignment score for each report $r_i \in \mathcal{B}$:
\begin{equation}
s_{\text{find}}(i)=
\frac{\mathbf{p}^{\top}\mathbf{y}_i}{\|\mathbf{p}\|_{1}+\epsilon}
\;-\;
\frac{(\mathbf{1}-\mathbf{p})^{\top}\mathbf{y}_i}{\|\mathbf{1}-\mathbf{p}\|_{1}+\epsilon}, \epsilon=10^{-8},
\end{equation}
which rewards agreement with high-probability findings and penalizes unlikely ones.
% This score follows radiologist reasoning as left-hand term ensures the inclusion of findings that the LQT predicts with high confidence. Conversely, the right-hand term penalizes exemplars containing findings that the LQT deems highly improbable. 
The final ranking score combines visual similarity and clinical alignment, where
% \begin{equation}
$s_i=1/2*(\cos(\mathbf{v},\mathbf{t}_i) + s_{\text{find}}(i))$.
% \end{equation} 
% \item \textbf{Coverage-aware greedy selection.}
To ensure the final $k$ exemplars collectively encompass all predicted pathologies, we apply a coverage-aware greedy selection.  
Let $T=\{c\mid p_c\ge0.5\}$ be the set of target findings with weights $w_c=\frac{p_c}{\sum_{j\in T}p_j}$,
we iteratively construct the set of examplars $\mathcal{R}$ by selecting:
\begin{equation}
i^*=\arg\max_{i\in\mathcal{P}\setminus \mathcal{R}}
\left[
s_i
+ \sum_{c\in T} w_c\,y_{i,c}\,\mathbf{1} \{c\notin\mathcal{C}\}
\right],
\end{equation}
where $\mathcal{C}$ tracks the labels already covered by previous selections, updating
$\mathcal{R}\leftarrow \mathcal{R}\cup\{i^*\}$ and
$\mathcal{C}\leftarrow \mathcal{C}\cup\{c\mid y_{i^*,c}=1\}$,
until $|\mathcal{R}|=k$.
% \end{enumerate}
% Given the predicted pathology labels from Stage 1, we retrieve the most clinically relevant reports from the training database. A lightweight projector, trained contrastively on paired CT scans and radiology reports, maps visual embeddings into a shared latent space with text, enabling retrieval ranked by cosine similarity. Crucially, this retrieval is also pathology-guided. 
% For a given scan, we first filter candidate reports to ensure their ground-truth labels align with the predicted findings from LQT, and then rank them by cosine similarity.
% This prevents the retrieval of generic templates when abnormalities are detected.
% At inference, we retrieve the top-$k$ reports from the training database. This pathology-guided retrieval provides the downstream LLM with both relevant clinical content and a realistic reporting style, grounded in actual radiology practice.

% We retrieve clinically relevant reports by combining learned visual–textual features with expert pathology labels. A projector trained contrastively with paired CT scans and reports maps images into the same latent space as text, enabling cosine similarity retrieval. This retrieval is guided by filtering candidates using "expert" pathology findings, ensuring stronger clinical alignment.

\paragraph{Stage 3: Frozen LLM synthesis.}
The final report is synthesized by a frozen medical LLM (MedGemma-27B \cite{sellergren2025medgemmatechnicalreport}), pretrained on large-scale medical corpora including radiology reports. The model receives as input: (i) the predicted pathology labels of LQT from Stage 1, (ii) the retrieved exemplar reports $\mathcal{R}$ from Stage 2, and (iii) available patient metadata (e.g., age, sex).
Conditioning on this structured clinical context, the LLM reasons over detected findings and their clinical relationships, draws on retrieved reports for appropriate terminology and reporting conventions, and generates a structured Findings section in realistic radiology style. It then produces a clinically consistent Impression by integrating the generated findings with its pretrained medical expertise.
The decision to keep the LLM frozen is deliberate. Fine-tuning on a small, imbalanced dataset risks degrading the medical reasoning and clinical language capabilities acquired during large-scale pretraining. A frozen LLM also avoids learning dataset-specific shortcuts that cause template collapse, as observed in fine-tuned VLMs. By restricting learning to Stages 1 and 2, CLarGen ensures that the LLM operates as a genuine medical reasoning module rather than a pattern-matching predictor.

% It jointly reasons over the pathologies, their clinical relationships and coherence together, analyses similar reports for present pathologies, their links ... but also for pathologies phrasing / style to generate a structured and fluent Findings section, following realistic radiology phrasing and style. The model then writes a clinically consistent Impression based on its generated findings, leveraging its pretrained medical knowledge. Keeping the LLM frozen preserves its reasoning capabilities while avoiding costly and unstable large-scale fine-tuning.

\vspace{-0.5em}
\section{Experiments}
\label{sec:experiments}
\vspace{-0.5em}
\paragraph{Dataset and preprocessing.} We evaluate on the CT-RATE dataset~\cite{hamamci2026generalist}, which comprises 25,678 chest CT studies with corresponding reports, using the official train/validation/test split.
For retrieval, the index is constructed exclusively from the training set; validation and test samples are retrieved only from this fixed index. To prevent information leakage, any report associated with the same patient as in the query is removed from the retrieval pool. 

% For the retrieval protocol the retrieval index is built only from the training split; validation samples retrieve exclusively from this index. We exclude any identical study/report from candidates to prevent leakage.

\paragraph{Models and Hyperparameters.}
We use a frozen 3D visual encoder HLIP~\cite{zhao2025towards}, and train only lightweight components: (i) a pathology-specific latent query Transformer head for multi-label classification with focal loss ($\gamma=2$), and (ii) a contrastive image-text projector used for retrieval. The medical LLM (MedGemma-27B~\cite{sellergren2025medgemmatechnicalreport}) is kept frozen and used only at inference time. All trained modules are optimized with AdamW (learning rate $6\times10^{-5}$, weight decay $0.008$) using a cosine annealing scheduler with warmup. We train for 20 epochs with an effective batch size of 64 in mixed precision (bf16). Retrieval uses cosine similarity in the joint embedding space with label-aware filtering, and we provide the top-$k$ retrieved reports to the LLM (default $k=5$). Unless specified otherwise, generation uses low-temperature decoding (temperature 0.2, top-$p$ 0.9, max 512 new tokens) for stable outputs. Experiments are run on one NVIDIA H100 GPU. Computational costs are reported in Sec.~\ref{sec:computational_cost}.

\paragraph{Evaluation Metrics.}
We evaluate across four complementary metric groups:

\noindent(i) \textbf{NLP metrics}
quantify linguistic quality via ROUGE-L, BLEU-4, and METEOR, measuring sequence-level and n-gram overlap with references.

\noindent(ii) \textbf{Clinical metrics}
evaluate medical correctness using macro-F1, precision, recall, and the Clinical Report Generation (CRG) score~\cite{hamamci2025crg}. Following \cite{hamamci2026generalist}, we use their pre-trained pathology classifier based on RadBERT-RoBERTa-4m~\cite{yan2022radbert}, to extract labels from generated text, ensuring an unbiased and standardized clinical comparison.
To quantify the dependence of detection on pathology prevalence, we additionally report \emph{prevalence-stratified recall}. Pathologies are split into quartiles by their test-set prevalence; $\mathrm{Surv}_{rare}$ is the mean recall over the bottom-prevalence quartile (rare findings) and $\mathrm{Surv}_{common}$ is the mean recall over the top quartile (common findings). A large gap $\Delta = \mathrm{Surv}_{common} - \mathrm{Surv}_{rare}$ signals that the model preserves common pathologies while dropping rare ones, the hallmark of template-driven generation.

\noindent(iii) \textbf{Diversity metrics}
diagnose lexical and semantic degeneration.
Distinct-1/2~\cite{li2016diversity} quantifies lexical diversity and detect repetitive phrasing patterns.
% Self-BLEU-4 for inter-report similarity,
We also report the effective $n$-gram vocabulary $\mathrm{effV}_n = 2^{H_n}$, where $H_n$ is the Shannon entropy of the corpus-level $n$-gram distribution. Unlike distinct-$n$, which counts unique types invariant to frequency, $\mathrm{effV}_n$ is \emph{frequency-weighted}: a corpus dominated by a handful of templates has a small $\mathrm{effV}_n$ even if its raw count of unique $n$-grams is large. Intuitively, it is the equivalent number of uniformly-used $n$-grams that would reproduce the observed entropy. We report $\mathrm{effV}_1$ (vocabulary scale) and $\mathrm{effV}_2$ (local phrase / template scale).
We also introduce a \textit{Normal Bias Similarity (NBS)} score inspired by~\cite{snell2017prototypical}, computing cosine similarity between each generated report and an embedding of normal-reference reports to detect semantic drift toward generic normal templates (details in Sec.~\ref{app:nbs}).
% Measure the ratio of unique unigrams and bigrams across generated reports. We use them to quantify lexical diversity and detect repetitive phrasing patterns typical of template-based generation.
% \item \textbf{Self-BLEU-4.}  
% Computes BLEU-4 between each generated report and the rest of the generated set. High self-BLEU indicates strong redundancy across outputs. We use this metric to explicitly measure inter-report similarity and identify mode collapse at the corpus level.

% \item \textbf{Normal Bias Similarity (NBS)}  
% Designed to capture semantic template collapse toward \emph{normal} findings. We first compute a reference embedding by averaging text embeddings of multiple ground-truth reports labeled as normal. We then measure the cosine similarity between this reference embedding and each generated report. High NBS values indicate that a model systematically produces reports semantically close to a generic normal template, even when pathology is present—an effect not reliably captured by surface-level diversity metrics.

\noindent(iv) \textbf{LLM-as-a-Judge}
provides a holistic evaluation using Gemini 2.5~\cite{comanici2025gemini}. Each generated report is scored against the reference on five criteria scoring from 0 to 10: factual clinical accuracy, global medical coherence, impression quality, radiology style/formalism, and clinical safety.

\paragraph{Baselines.}
We compare with existing publicly available open-source methods including CT-CHAT \cite{hamamci2026generalist}, SAMF \cite{hosseini2025samf}, Reg2Rg \cite{chen2025largelanguagemodelregionguided}, and HULU-med \cite{jiang2025hulu}.
CT-CHAT is a vision-language chat model for 3D chest CT volumes, built by coupling CT-CLIP's vision encoder with a pretrained LLM fine-tuned on millions of question-answer pairs. SAMF fuses 2D slice-level and 3D volumetric features across multiple scales. Reg2RG enhances CT report generation by grounding each finding in specific anatomical regions. HULU-Med is a generalist medical VLM that supports multiple input modalities across a broad range of clinical tasks.
To ensure reproducibility and facilitate future benchmarking, we will release the raw outputs for all models alongside our codebase.

% We also attempted to re-train CT2Rep \cite{hamamci2024ct2rep}, but failed to converge and produced very poor results; therefore, it is not included in our comparisons. 
% To the best of our knowledge, we include all available 3D report generation models with publicly shared weights. 
% All reports generated by every evaluated method, including ours, will be released as JSON files to ensure full reproducibility and allow re-computation of all metrics. 
% For CT2Rep only, we report the results provided by the original authors, as we were unable to have a good training, unlike all other baselines which are fully re-evaluated by us.

\section{Results and Discussion}
\label{sec:results}

\begin{table*}[t]
\centering

\caption{
Comprehensive comparison across four complementary dimensions:
(a) \emph{NLP fidelity}, measuring lexical overlap with reference reports;
(b) \emph{clinical correctness}, evaluating pathology consistency and clinical relevance;
(c) \emph{diversity and template conformity}, quantifying lexical richness and tendency toward template generation;
and (d) \emph{prevalence-stratified survival}, assessing robustness on rare versus common pathologies.
$\mathrm{Surv}_{rare}$ and $\mathrm{Surv}_{common}$ denote recall on the rarest and most common pathology quartiles, respectively, and $\Delta$ measures the performance gap between them.
Values are reported across three runs unless otherwise noted. (CRG = Clinical Report Generation score;
NBS = Normal Bias Similarity;
$\mathrm{effV}_n = 2^{H_n}$ denotes effective $n$-gram vocabulary size (single run, deterministic given the model output corpus).)
}

\label{tab:all_results}

\small
\setlength{\tabcolsep}{6pt}
\renewcommand{\arraystretch}{1.08}

% =========================================================
% TOP PANELS
% =========================================================
\begin{tabular}{@{}lccc|cccc@{}}
\toprule

\multicolumn{4}{c|}{\textbf{(a) NLP fidelity}}
&
\multicolumn{4}{c}{\textbf{(b) Clinical correctness}}
\\

\cmidrule(lr){1-4}
\cmidrule(lr){5-8}

\textbf{Method}
& ROUGE-L$\uparrow$
& BLEU$\uparrow$
& METEOR$\uparrow$
&
F1$\uparrow$
& Precision$\uparrow$
& Recall$\uparrow$
& CRG$\uparrow$
\\

\midrule

CT-CHAT
& \sd{0.382}{0.004}
& \bsd{0.277}{0.004}
& \bsd{0.399}{0.003}
&
\sd{0.189}{0.010}
& \bsd{0.493}{0.018}
& \sd{0.164}{0.012}
& \sd{0.368}{0.008}
\\

HULU-Med
& \sd{0.340}{0.006}
& \sd{0.239}{0.005}
& \sd{0.374}{0.004}
&
\sd{0.194}{0.012}
& \sd{0.303}{0.015}
& \sd{0.166}{0.012}
& \sd{0.365}{0.009}
\\

Reg2Rg
& \sd{0.245}{0.007}
& \sd{0.275}{0.004}
& \sd{0.368}{0.004}
&
\sd{0.185}{0.011}
& \sd{0.373}{0.017}
& \sd{0.138}{0.011}
& \sd{0.363}{0.010}
\\

SAMF
& \bsd{0.388}{0.003}
& \sd{0.269}{0.004}
& \sd{0.388}{0.004}
&
\sd{0.097}{0.009}
& \sd{0.144}{0.011}
& \sd{0.086}{0.008}
& \sd{0.355}{0.009}
\\

\midrule

\textbf{CLarGen}
& \sd{0.298}{0.005}
& \sd{0.208}{0.005}
& \sd{0.335}{0.005}
&
\bsd{0.486}{0.012}
& \sd{0.405}{0.015}
& \bsd{0.646}{0.014}
& \bsd{0.472}{0.010}
\\

\bottomrule
\end{tabular}

\vspace{3.5mm}

% =========================================================
% BOTTOM PANELS
% =========================================================
\begin{tabular}{@{}lccccc|ccc@{}}
\toprule

\multicolumn{6}{c|}{\textbf{(c) Diversity and template conformity}}
&
\multicolumn{3}{c}{\textbf{(d) Prevalence-stratified survival}}
\\

\cmidrule(lr){1-6}
\cmidrule(lr){7-9}

\textbf{Method}
& Distinct-1$\uparrow$
& effV$_1$$\uparrow$
& Distinct-2$\uparrow$
& effV$_2$$\uparrow$
& NBS$\downarrow$
&
$\mathrm{Surv}_{rare}$$\uparrow$
& $\mathrm{Surv}_{common}$$\uparrow$
& $\Delta$$\downarrow$
\\

\midrule

\rowcolor{gray!12}
Ground truth
& 0.016
& 296
& 0.094
& 2402
& --
&
1.000
& 0.999
& -0.001
\\

% \midrule

CT-CHAT
& \sd{0.002}{0.0002}
& \nostd{142}
& \sd{0.006}{0.0005}
& \nostd{392}
& \sd{0.877}{0.010}
&
\nostd{0.035}
& \nostd{0.207}
& \nostd{0.173}
\\

HULU-Med
& \bsd{0.014}{0.0008}
& \underline{\nostd{230}}
& \bsd{0.080}{0.0030}
& \underline{\nostd{1393}}
& \underline{\sd{0.845}{0.012}}
&
\nostd{0.058}
& \underline{\nostd{0.300}}
& \nostd{0.241}
\\

Reg2Rg
& \sd{0.004}{0.0004}
& \nostd{151}
& \sd{0.018}{0.0012}
& \nostd{528}
& \sd{0.923}{0.009}
&
\underline{\nostd{0.088}}
& \nostd{0.195}
& \underline{\nostd{0.107}}
\\

SAMF
& \sd{0.001}{0.0001}
& \nostd{114}
& \sd{0.002}{0.0002}
& \nostd{271}
& \sd{0.918}{0.010}
&
\nostd{0.000}
& \nostd{0.187}
& \nostd{0.187}
\\

\midrule

\textbf{CLarGen}
& \underline{\sd{0.011}{0.0007}}
& \textbf{\nostd{279}}
& \underline{\sd{0.054}{0.0025}}
& \textbf{\nostd{1826}}
& \bsd{0.730}{0.015}
&
\textbf{\nostd{0.628}}
& \textbf{\nostd{0.622}}
& \textbf{\nostd{-0.006}}
\\

\bottomrule
\end{tabular}

% \vspace{1.5mm}

% \footnotesize{
% \textbf{Abbreviations:}
% CRG = clinical report generation score;
% NBS = normalized boilerplate score;
% $\mathrm{effV}_n = 2^{H_n}$ denotes effective $n$-gram vocabulary size.
% }

\end{table*}

\begin{figure*}[h]
\centering
\begin{tikzpicture}
    \begin{axis}[
        ybar,
        bar width=7pt,
        width=\textwidth,
        height=3.5cm,
        ymax=8.2, ymin=3,
        ylabel={Score (1-10)},
        symbolic x coords={Factual Accuracy, Medical Coherence, Impression Quality, Radiology Style, Clinical Safety},
        xtick=data,
        xticklabel style={rotate=0, font=\scriptsize, align=center},
        nodes near coords={}, 
        legend style={at={(0.5,1.15)}, anchor=north, legend columns=-1, draw=none, font=\scriptsize},
        ymajorgrids=true,
        grid style=dashed,
        enlarge x limits=0.15,
    ]

    % CT-CHAT [5.67, 5.29, 4.35, 7.35, 5.67]
    \addplot[fill=bline1, draw=black!50] coordinates {
        (Factual Accuracy,5.67) (Medical Coherence,5.29) (Impression Quality,4.35) (Radiology Style,7.35) (Clinical Safety,5.67)
    };
    
    % HULU [5.62, 4.95, 4.05, 6.40, 5.08]
    \addplot[fill=bline2, draw=black!50] coordinates {
        (Factual Accuracy,5.62) (Medical Coherence,4.95) (Impression Quality,4.05) (Radiology Style,6.40) (Clinical Safety,5.08)
    };

    % Reg2Rg [5.51, 4.85, 3.84, 7.01, 5.12]
    \addplot[fill=bline3, draw=black!50] coordinates {
        (Factual Accuracy,5.51) (Medical Coherence,4.85) (Impression Quality,3.84) (Radiology Style,7.01) (Clinical Safety,5.12)
    };

    % SAMF [4.90, 4.27, 3.34, 6.84, 4.49]
    \addplot[fill=bline4, draw=black!50] coordinates {
        (Factual Accuracy,4.90) (Medical Coherence,4.27) (Impression Quality,3.34) (Radiology Style,6.84) (Clinical Safety,4.49)
    };

    % CLarGen (MedGemma data) [7.82, 7.12, 6.27, 7.11, 7.04]
    \addplot[fill=ourblue, draw=black, postaction={pattern=north east lines, pattern color=white!30}] coordinates {
        (Factual Accuracy,7.42) (Medical Coherence,7.12) (Impression Quality,6.27) (Radiology Style,6.74) (Clinical Safety,7.04)
    };

    \legend{CT-CHAT, HULU, Reg2Rg, SAMF, \textbf{CLarGen}}
    \end{axis}
\end{tikzpicture}

\vspace{-0.5em}
\caption{\textbf{LLM-as-a-judge Evaluation.} VLM baselines (grey bars) score highly on Radiology Style but drop sharply in clinical evaluation, while CLarGen (blue bars) remains high across all metrics.}
\label{fig:qualitative_metrics}
\vspace{-0.6em}
\end{figure*}
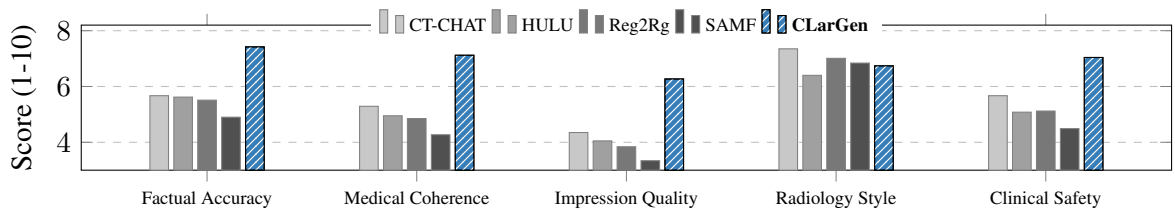

\subsection{Systematic Analysis on Template Collapse}
Our evaluation reveals a critical disconnect between linguistic quality and clinical utility of current VLM paradigms.
As shown in Table~\ref{tab:all_results}, while all VLM baselines achieve high NLP scores and radiology formalism, their clinical performance is severely limited, with macro-F1 scores below 0.20 and low factual accuracy.
The diversity analysis in Table~\ref{tab:all_results}(c) reveals that this discrepancy is a direct result of \emph{template collapse}.
Most existing VLM models exhibit low Distinct-1/2 scores and high NBS scores, indicating that the models have converged to a singular, generic template with restricted vocabulary and high inter-report similarity.
This collapse is quantified by the effective $n$-gram vocabulary $\mathrm{effV}_n$ (Table~\ref{tab:all_results}(c)): at the bigram level, VLM baselines occupy only 271--1393 effective bigrams, an order of magnitude below the 2402 of ground-truth reports; SAMF in particular has $\mathrm{effV}_2 = 271$, meaning its entire test-set output is statistically equivalent to a corpus circulating fewer than 300 distinct phrases. CLarGen, by contrast, reaches 76\% of the ground-truth effective bigram repertoire, confirming that decoupling perception from generation preserves both lexical breadth and usage balance.
Fig.~\ref{fig:hook_templates_rarity} visualizes this in the latent space, where VLM baselines form tight clusters that fails to capture the variance of the ground-truth report distribution.
Table~\ref{tab:dominant_templates} makes this concrete at the 
case level: CT-CHAT emits a single normal-template report for 
28.9\% of all test cases, and SAMF for 20.8\%, compared to 
only 0.8\% in the ground truth and 1.8\% in CLarGen.
% Bailiang's version 05/26
Appendix Table~\ref{tab:tc_diagnostics} shows the same pattern: baseline outputs require far fewer normlized templates to cover most reports, whereas CLarGen remains closer to the ground-truth distribution.
HULU-Med~\cite{jiang2025hulu} preserves high distinct-1/2 among baselines, yet its $\mathrm{effV}_1$ and $\mathrm{effV}_2$ remain below those of CLarGen, and suffers from low clinical accuracy.
This degeneration coincides with low recall and comparatively higher precision, suggesting a bias toward predicting normal findings while failing to report present abnormalities.
We hypothesize that under the constraints of extreme class imbalance and noisy 3D signals, next-token prediction incentivizes the model to ignore weak visual evidence in favor of the most statistically frequent textual patterns. Further analyses see Sec.~\ref{app:template_collapse_evidence}.

% All VLM variants achieve high NLP metrics, consistent with published literature. However, clinical performance remains limited (Table~\ref{tab:main_results}). Notably, our systematic architectural improvements do not substantially address this gap.
% This behavior correlates with a tendency to produce repetitive "normal" reports. Diversity metrics (Table~\ref{tab:diversity}) quantify this: VLMs achieve low distinct-1 scores, indicating restricted vocabulary, while self-BLEU-4 is high, meaning generated reports are more identical to each other. 
% Figure~\ref{fig:embedding_space} visualizes this collapse: CT-CHAT outputs cluster tightly in embedding space, contrasting with the diverse distribution of ground truth reports.
% This behavior may arise from the interaction between training objective and data characteristics. VLMs optimized via next-token prediction can gravitate toward outputs that minimize loss on the majority of cases. Under CT-RATE's extreme class imbalance and limited 3D training data, models may converge to "normal" templates rather than learning to detect rare pathologies from noisy visual representations.

%%%% BEGINNING
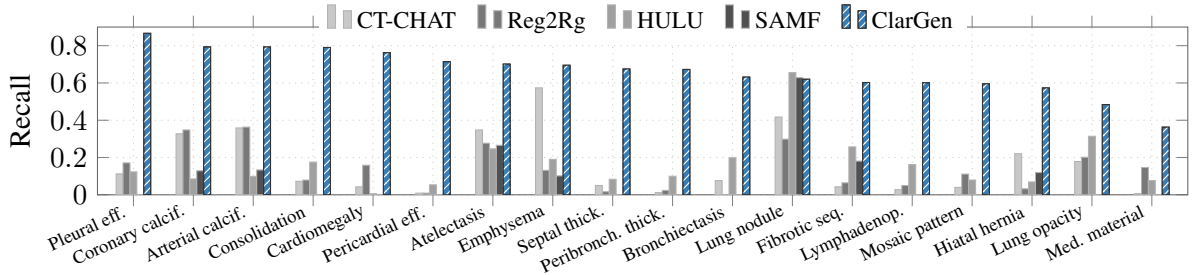
\begin{figure*}[t]
\centering
\begin{tikzpicture}
\pgfplotstableread[col sep=comma]{
Pathology,CT-CHAT,Reg2Rg,HULU,SAMF,ClarGen
Pleural eff.,0.112299,0.171123,0.122995,0.000000,0.866310
Coronary calcif.,0.327320,0.347938,0.085052,0.128866,0.793814
Arterial calcif.,0.358621,0.363218,0.098851,0.133333,0.793103
Consolidation,0.072165,0.079038,0.175258,0.000000,0.790378
Cardiomegaly,0.042683,0.158537,0.006098,0.000000,0.762195
Pericardial eff.,0.008929,0.008929,0.053571,0.000000,0.714286
Atelectasis,0.347945,0.276712,0.246575,0.265753,0.701370
Emphysema,0.573770,0.131148,0.190164,0.101639,0.695082
Septal thick.,0.050000,0.016667,0.083333,0.000000,0.675000
Peribronch. thick.,0.011696,0.023392,0.099415,0.000000,0.672515
Bronchiectasis,0.076023,0.000000,0.198830,0.000000,0.631579
Lung nodule,0.417143,0.298571,0.655714,0.628571,0.620000
Fibrotic seq.,0.042857,0.064286,0.257143,0.180952,0.602381
Lymphadenop.,0.027431,0.049875,0.162095,0.000000,0.600998
Mosaic pattern,0.039683,0.111111,0.079365,0.000000,0.595238
Hiatal hernia,0.220183,0.032110,0.068807,0.119266,0.573394
Lung opacity,0.178512,0.201653,0.314050,0.000000,0.484298
Med. material,0.006369,0.146497,0.076433,0.000000,0.363057
}\datatable

\begin{axis}[
    ybar=0pt, 
    bar width=2.6pt, 
    width=\textwidth,
    height=3.8cm, 
    ymin=0, ymax=0.9, 
    ylabel={Recall},
    xtick={0,...,17},
    xticklabels from table={\datatable}{Pathology},
    x tick label style={rotate=25, anchor=east, font=\scriptsize, align=right},
    enlarge x limits={abs=0.6},
    nodes near coords={}, 
    grid=both,
    major grid style={dotted, gray!40},
    axis line style={draw=black!60},
    legend style={
        at={(0.5,0.9)}, 
        anchor=south,
        legend columns=-1, 
        /tikz/every even column/.append style={column sep=6pt},
        draw=none, 
        font=\small
    },
]
\addplot[fill=bline1, draw=black!30] table[x expr=\coordindex, y=CT-CHAT]{\datatable};
\addplot[fill=bline3, draw=black!30] table[x expr=\coordindex, y=Reg2Rg]{\datatable};
\addplot[fill=bline2, draw=black!30] table[x expr=\coordindex, y=HULU]{\datatable};
\addplot[fill=bline4, draw=black!30] table[x expr=\coordindex, y=SAMF]{\datatable};
\addplot[fill=ourblue, draw=black!80, postaction={pattern=north east lines, pattern color=white!40}] table[x expr=\coordindex, y=ClarGen]{\datatable};

\legend{CT-CHAT, Reg2Rg, HULU, SAMF, ClarGen}
\end{axis}

\end{tikzpicture}
\vspace{-1.em}
\caption{Recall by pathology of generated reports from different methods.}
\label{fig:pathology_recall}
\vspace{-1.em}
\end{figure*}

%%%% END 

 % \begin{table}[h]
 %    \centering
 %    \caption{\textbf{Prevalence-stratified survival.} Recall on the rarest- vs.\ most-common-quartile pathologies, computed from RadBERT-extracted labels on the test set. $\Delta$ is the gap between common and rare.}
 %    \label{tab:survival}
 %    \small
 %    \setlength{\tabcolsep}{6pt}
 %    \begin{tabular}{@{}lccc@{}}
 %      \toprule
 %      \textbf{Model} & $\mathrm{Surv}_{rare}\uparrow$
 %                     & $\mathrm{Surv}_{common}\uparrow$
 %                     & $\Delta\downarrow$ \\
 %      \midrule
 %      Ground truth & 1.000 & 0.999 & -0.001 \\
 %      CT-CHAT      & \nostd{0.035} & \nostd{0.207} & \nostd{0.173} \\
 %      HULU-Med     & \nostd{0.058} & \underline{\nostd{0.300}} & \nostd{0.241} \\
 %      Reg2Rg       & \underline{\nostd{0.088}} & \nostd{0.195} & \underline{\nostd{0.107}} \\
 %      SAMF         & \nostd{0.000} & \nostd{0.187} & \nostd{0.187} \\
 %      \midrule
 %      \textbf{CLarGen}
 %                   & \textbf{\nostd{0.628}} & \textbf{\nostd{0.622}} & \textbf{\nostd{-0.006}} \\
 %      \bottomrule
 %    \end{tabular}
 %  \end{table}

% \vspace{-0.6em}

% \subsection{Sensitivity to Rare Pathology Detection}
% Bailiang's version
\subsection{Template Collapse Suppresses Rare Findings Detection}
Rare-finding suppression is the most clinically critical consequence of Template Collapse.
Table~\ref{tab:all_results}(b) and Fig.~\ref{fig:pathology_recall} show that baselines yield high precision but low recall, with near-zero recall for many infrequent findings. This indicates a conservative failure mode, where the models maintain accuracy by generating generic normal templates and avoiding the risk of reporting abnormalities.
% This preserves NLP fluency but systematically omits rare yet critical pathologies.
The rare and common pathology detection decomposition in Table~\ref{tab:all_results}(d) makes the failure mode explicit.
Across all VLM baselines, recall on the rare-pathology quartile collapses to between $0.000$ (SAMF) and $0.088$ (Reg2Rg), while recall on the common-pathology quartile remains an order of magnitude higher (0.19-0.30). The resulting common$-$rare gap $\Delta$ ranges from $0.11$ to $0.24$.
This is the conservative-template signature: by defaulting to common ``normal'' descriptions and avoiding rare abnormalities, baselines preserve NLP fluency at the cost of clinical accuracy. 
In contrast, CLarGen achieves substantially higher recall across all findings while maintaining high precision. It achieves $\mathrm{Surv}_{rare} = 0.628$ vs.\ $\mathrm{Surv}_{common} = 0.622$, recovering rare findings at essentially the same rate as common ones, indicating a balanced diagnostic ability.
This behavior stems from the explicit decoupling of clinical perception and language generation, which forces the model to attend to weak visual signals that joint pipelines overlook, essential for robust report generation. \\

% Other methods exhibit a very low recall (see Table \ref{tab:main_results}). This means that they miss most of the pathologies that are actually present. When they make a statement on normal reports, it is often correct (higher precision), but they almost never report anything abnormal. This is the signature of a model that collapses toward predicting “normal”: conservative, silent about anomalies, and therefore dangerous in a clinical setting.
% In contrast, our method achieves a substantially higher recall, while maintaining a comparable level of precision.

\vspace{-0.6em}

% \subsection{CLarGen's High Clinical Performance}
% Bailiang's version
\subsection{CLarGen Mitigates Collapse and Improves Clinical Accuracy}
CLarGen mitigates Template Collapse by preserving clinical evidence during generation. It achieves a macro-F1 of 0.486 and a CRG of 0.472, outperforming all baselines (Table \ref{tab:all_results}b) and achieving nearly a 150\% improvement in clinical detection accuracy over the best-performing VLM baseline (HULU-Med, F1 = 0.194).
Notably, CLarGen exhibits a principled trade-off: while its NLP scores are slightly lower, this indicates reduced lexical bias. By avoiding template collapse, CLarGen produces diverse, patient-specific reports that naturally have lower $n$-gram overlap with the generic normal reports that dominate the test set. 
This is indicated by its substantially more diverse outputs, shown as high diversity metrics (Table~\ref{tab:all_results}c) and widespread latent space matching the variance of the ground-truth report distribution (Fig.~\ref{fig:hook_templates_rarity}).
Crucially, its NLP scores remain sufficiently high to confirm that outputs are fluent and well-structured, striking an optimal balance between clinical accuracy and linguistic fluency. 
The LLM-as-judge evaluation in Fig.~\ref{fig:qualitative_metrics} provides independent assessment from a holistic, reference-free
perspective. Baselines score highly on \textit{Radiology Style}, confirming they have learned the surface form of radiology reports, but drop sharply on \textit{Factual Accuracy}, \textit{Medical Coherence}, and \textit{Clinical Safety}: dimensions most relevant to clinical deployment. By contrast, CLarGen leads on all clinical aspects while remaining competitive on Radiology Style. This convergent evidence across automated classifiers and LLM judgment confirms CLarGen's robust clinical gains.
\\
% As illustrated in Figure~\ref{fig:metric_tradeoff}, CLarGen breaks the inverse correlation between NLP fluency and F1-score, striking an optimal balance between linguistic coherence and diagnostic specificity.

% As shown in Table \ref{tab:main_results}, CLarGen achieves substantially improved clinical metrics  with an F1-score of 0.486 and a CRG score of 0.472, decisively outperforming all VLM baselines, demonstrating robust pathology detection, particularly for rare findings that VLMs systematically miss.
% CLarGen obtains moderately lower NLP metrics than CT-CHAT. However, this represents a principled trade-off: avoiding template collapse naturally results in lower lexical overlap with any single reference report, as confirmed by its substantially more diverse outputs (Table \ref{tab:diversity} and Figure \ref{fig:tsne}). Its NLP scores remain high enough to demonstrate coherent, well-structured reports with appropriate medical terminology, but are not excessively high—which, as VLM results show, would indicate template overfitting (\ref{fig:metric_tradeoff}). CLarGen strikes the optimal balance: clinically accurate and linguistically fluent, without sacrificing specificity for statistical conformity, making it more suitable for clinical deployment where pathology detection accuracy is paramount.

\vspace{-0.8em}

\subsection{Impact of Visual Encoder Quality}
\label{sec:encoder_quality}
A key advantage of decoupling classification from generation is that improvements in visual representation directly translate to better report quality. 
We evaluate three encoders of increasing strength: CT-CLIP~\cite{hamamci2026generalist}, Curia~\cite{dancette2025curiamultimodalfoundationmodel}, and HLIP~\cite{zhao2025towards}, under three settings: (i)~a linear classifier on frozen embeddings, measuring raw detection potential; (ii)~a CT-CHAT-style LLaVA VLM fine-tuned end-to-end; and (iii)~our CLarGen pipeline.

As shown in Table~\ref{tab:encoder_ablation}, the classifier-only upper bound improves substantially from 0.385 to 0.512 as encoder quality increases, confirming that stronger encoders carry richer pathology signals. The end-to-end VLM, however, fails to exploit this: its F1 remains nearly flat (0.184-0.193), indicating that the generation objective itself becomes the bottleneck. In contrast, CLarGen scales faithfully with encoder quality, rising from 0.342 to 0.486 and recovering 89\%-95\% of the raw detection potential. Note that this remaining gap reflects evaluation noise rather than information loss: report-level F1 requires re-extracting labels from generated text via RadBERT, whose own classification errors impose an upper bound on any text-based metric. The near-complete recovery achieved by CLarGen therefore suggests that encoder improvements propagate almost losslessly through our pipeline, making it straightforward to benefit from future advances in medical vision encoders. \\
\begin{table}[t]
    \centering
    \small
    % \caption{\textbf{Impact of Visual Encoder Quality.} We compare the raw detection potential (classification of the same 18 pathologies via a trained classifier head on top of visual encoder embeddings) against the final generated report performance (classification via a private text classifier on the same 18 pathologies). While end-to-end VLMs stagnate regardless of the encoder quality, \textbf{CLarGen} effectively scales with better visual representations, closely matching the raw classification performance. All VLM variants follow CT-CHAT training.}
    \caption{Impact of Visual Encoder Quality. We compare the raw detection potential against the final generated report performance for different visual encoders.}
    \vspace{-0.8em}
    \label{tab:encoder_ablation}
    \setlength{\tabcolsep}{4pt} 
    \renewcommand{\arraystretch}{0.9}
    \begin{tabular}{@{} l c c c @{}}
    \toprule
    & Potential & \multicolumn{2}{c}{Report Generation F1} \\
    \cmidrule(lr){2-2} \cmidrule(l){3-4}
    Visual Encoder & \begin{tabular}{@{}c@{}}Classifier  \\ Only\end{tabular} & \begin{tabular}{@{}c@{}}Conventional \\ VLM\end{tabular} & \begin{tabular}{@{}c@{}}CLarGen \\ (Ours)\end{tabular} \\
    \midrule
    CT-CLIP & 0.385 & 0.189 & 0.342 \\
    Curia & 0.464 & 0.184 & 0.431 \\ 
    HLIP  & \textbf{0.512} & \textbf{0.193} & \textbf{0.486} \\
    \bottomrule
    \end{tabular}
    % \vspace{-1.8em}
\end{table}

\vspace{-0.6em}

\subsection{Robustness to Label Extractors}
To verify that our clinical evaluation does not depend on RadBERT-specific behavior, we re-extracted the 18 labels using three independently developed model families: Qwen3-30B-A3B-Instruct, Gemini-flash-lite, and Claude Opus~4.8. Each extractor was applied under identical label definitions to the same validation reports and their corresponding CLarGen and CT-CHAT generations; macro-F1 was computed per extractor against the labels it independently assigned to the matching ground-truth report, yielding a fully matched comparison. Results in Table~\ref{tab:extractor_robustness} show that CLarGen's clinical F1 remains stable at 0.49--0.51 across all four extractors and consistently exceeds CT-CHAT, preserving a gap of 0.28--0.32 in every case; these values closely reproduce the paper's full-test-set results (0.487 vs.\ 0.189), indicating that CLarGen's clinical performance gain does not depend on RadBERT.
We further assessed how consistently the four extractors label the same 60 shared ground-truth reports, as reported in Table~\ref{tab:extractor_agreement}. RadBERT agrees closely with each of the three independently developed model families on the same ground-truth reports (98.2--98.5\% cell agreement; $\kappa=0.934$--$0.947$). We treat these models as independent sensitivity checks rather than replacements for RadBERT. 
% combined with RadBERT's prior validation against radiologist annotations, this agreement indicates that its outputs do not rely on any single extractor. 
Taken together, these results demonstrate that the clinical evaluation is not driven by RadBERT-specific behavior and that CLarGen's clinical performance is robust across alternative extractors.

\begin{table}[t]
\centering
\small
\caption{CLarGen's clinical F1 score gain is preserved under different label extractors.}
\label{tab:extractor_robustness}
\begin{tabular}{lccc}
\toprule
Label extractor &  CLarGen & CT-CHAT & Gain \\
\midrule
RadBERT       & 0.493 & 0.206 & +0.287 \\
Qwen3-30B-Instruct    & 0.509 & 0.222 & +0.287 \\
Gemini-flash-lite       & 0.498 & 0.218 & +0.280 \\
Claude Opus~4.8         & 0.495 & 0.172 & +0.323 \\
\bottomrule
\end{tabular}
\end{table}

\begin{table}[t]
\centering
\small
\caption{RadBERT labels are closely reproduced by independent extractors on the GT reports.}
\label{tab:extractor_agreement}
\begin{tabular}{lcc}
\toprule
Pair & Cell agreement & Cohen's $\kappa$ \\
\midrule
RadBERT vs.\ Qwen    & 98.5\% & 0.947 \\
RadBERT vs.\ Gemini  & 98.2\% & 0.934 \\
RadBERT vs.\ Claude  & 98.2\% & 0.936 \\
\bottomrule
\end{tabular}
\end{table}

\subsection{Ablation Studies}
% \vspace{-0.2em}
\textbf{(A) Retrieval strategy:} We evaluated the contribution of pathology-guided retrieval by comparing CLarGen with two standard retrieval-augmented generation (RAG) baselines using the same frozen LLM:
(i) visual embedding similarity (Vol$\leftrightarrow$Vol) which retrieves K nearest reports based on cosine similarity in the visual embedding space, and (ii) joint image-text similarity (Vol$\leftrightarrow$Text) based on the cosine similarity between the input visual embedding and text embeddings in a joint contrastive latent space.
As shown in Table~\ref{tab:rag}, these baselines achieve high NLP metrics but lower clinical accuracy,
indicating that embedding-based retrieval often prioritizes stylistic and lexical proximity over diagnostic equivalence. By contrast, CLarGen’s pathology-guided filtering acts as a semantic judge that ensures retrieved exemplars share the predicted diagnostic profile. It also compensates for the inherent noise brought by 3D foundation encoders, ensuring the LLM receives contextually relevant clinical evidence. These results validate that explicit clinical supervision at the retrieval stage is essential to ground generation, especially for limited and imbalanced 3D medical imaging. \\
\textbf{(B) Number of retrieved reports ($k$):} F1-score increases from 0.483 ($k=4$) to 0.489 ($k=5$) and then shows only marginal gains (0.490–0.495 for $k\geq6$). We therefore select $k=5$, which achieves near-optimal clinical performance while reducing prompt length and redundancy. \\
% \textbf{(C) Visual Encoder:} We evaluated different 3D visual encoders including CT-CLIP~\cite{hamamci2026generalist}, Curia~\cite{dancette2025curiamultimodalfoundationmodel}, and HLIP~\cite{zhao2025towards}, where HLIP leads to substantially higher clinical accuracy (F1-score: 0.486) compared to Curia (0.431) and CT-CLIP (0.342).
\textbf{(C) LLM input:} 
The Image-only setup follows a LLaVA-style design with projected HLIP visual tokens and LoRA fine-tuning of the LLM. It achieves reasonable NLP scores but weak clinical accuracy, suggesting limited pathology grounding from direct visual conditioning (more analyses in Sec.~\ref{app:sec:image-input-ablation}). Labels alone perform poorly, while exemplars substantially improve performance. Adding labels to exemplars further strengthens clinical accuracy.\\
% The \textit{Image only} variant follows a LLaVA-style setup: HLIP visual tokens are mapped through a trainable projector and prepended to the LLM input, while the LLM is fine-tuned with LoRA adapters. 
% This variant achieves high NLP scores but substantially lower clinical accuracy than exemplar-based conditioning, suggesting that direct visual-token conditioning mainly recovers plausible reporting style but remains insufficient for reliable pathology grounding in 3D CT. 
% Labels alone yield low NLP scores, confirming that fluent generation requires structural grounding from retrieved exemplars. Exemplars alone achieve strong performance, while adding labels maintains clinical accuracy and explicitly anchors generation to detected findings.\\
\textbf{(D) Base LLMs:} Table~\ref{tab:rag} also shows that using a domain-specific LLM (MedGemma 27B~\cite{sellergren2025medgemmatechnicalreport}) improves both clinical and NLP metrics than a general-domain LLM (LLaMA 3.1 8B~\cite{grattafiori2024llama}), indicating that a domain-specific backbone is better equipped to synthesize the LQT’s clinical signals into grounded diagnostic reasoning. 
\begin{table}[t]
\small
  \centering
  \setlength{\tabcolsep}{3pt}
  \renewcommand{\arraystretch}{0.9}
  \caption{Retrieval and LLM conditioning ablations.}
  \vspace{-0.8em}
  \label{tab:rag}
  % \begin{tabular}{@{}llcccc@{}}
  %   \toprule
  %   & & ROUGE-L $\uparrow$ & BLEU $\uparrow$ & F1 $\uparrow$ & CRG $\uparrow$ \\
  %   \midrule
  %   \multirow{3}{*}{Retrival} & Vol$\leftrightarrow$Vol & 0.330 & 0.237 & 0.393 & 0.422 \\
  %   & Vol$\leftrightarrow$Text & \textbf{0.347} & \textbf{0.263} & 0.446 & 0.439 \\
  %   & Pathology-guided Retrival & 0.298 & 0.208 & \textbf{0.486} & \textbf{0.472} \\
  %   \midrule
  %   \multirow{2}{*}{Base LLM} & LLaMA 3.1 8B & 0.276 & 0.181 & 0.472 & 0.461 \\
  %   & MedGemma 27B    & \textbf{0.298} & \textbf{0.208} & \textbf{0.486} & \textbf{0.472} \\
  %   \bottomrule
  % \end{tabular}
  \begin{tabular}{@{}lcccc@{}}
    \toprule
     & ROUGE-L $\uparrow$ & BLEU $\uparrow$ & F1 $\uparrow$ & CRG $\uparrow$ \\
    \midrule
    \textit{Retrieval} & \\\midrule
    Vol$\leftrightarrow$Vol & 0.330 & 0.237 & 0.393 & 0.422 \\
    Vol$\leftrightarrow$Text & \textbf{0.347} & \textbf{0.263} & 0.446 & 0.439 \\
    Pathology-guided & 0.298 & 0.208 & \textbf{0.486} & \textbf{0.472} \\
    \midrule
    \textit{LLM Input} & \\\midrule
    % Labels only    & 0.086 & 0.007 & 0.351 & 0.376 \\
    % % Image only & \\
    % Exemplars only & 0.342 & 0.259 & 0.499 & 0.484 \\
    % % Labels + Exemplars & \\
    % % Labels + Exe. + Img. & \\
    % Labels + Exemplars & 0.310 & 0.210 & 0.497 & 0.477 \\
    Labels only    & 0.120 & 0.002 & 0.408 & 0.449 \\

    Image only & \textbf{0.385} & \textbf{0.329} & 0.294 & 0.397\\ 
    % Image only & \\
    Exemplars only & 0.324 & 0.190 & 0.417 & 0.440 \\
    % Labels + Exemplars & \\
    % Labels + Exe. + Img. & \\
    Labels + Exemplars & 0.310 & 0.210 & \textbf{0.497} & \textbf{0.477} \\
    \midrule
    \textit{Base LLM} & \\\midrule
    LLaMA 3.1 8B & 0.276 & 0.181 & 0.472 & 0.461 \\
    MedGemma 27B    & \textbf{0.298} & \textbf{0.208} & \textbf{0.486} & \textbf{0.472} \\
    \bottomrule
  \end{tabular}
  \vspace{-0.5em}
\end{table}

\section{Conclusion}
We identified as a measurable failure mode in 3D CT report generation: autoregressive VLMs produce fluent radiology-style reports while suppressing patient-specific abnormalities.
Our analysis shows that this failure appears across clinical accuracy, output diversity, normal-template bias, dominant-template concentration, and rare-finding survival.
These metrics expose a mismatch that surface-overlap scores obscure: reports can resemble references lexically while omitting clinically relevant findings.
CLarGen mitigates this failure by separating clinical perception from language synthesis.
A Latent Query Transformer predicts pathology evidence, pathology-guided retrieval supplies clinically matched exemplars, and a frozen medical LLM converts this structured evidence into a report.
This design improves macro-F1 and rare-finding survival while preserving realistic reporting style.
Additional image-input VLM analyses show that direct visual conditioning alone does not solve collapse; autoregressive decoders still suppress weak clinical evidence unless clinical content is made explicit before generation.
These findings suggest that reliable 3D CT report generation requires explicit clinical grounding rather than stronger text generation alone.
Future work should improve upstream detection, validate the framework beyond CT-RATE, and study how generated reports should communicate uncertainty-aware clinical labels safely.

% \clearpage

\section*{Limitations}

While CLarGen improves clinical grounding and reduces template collapse, several limitations remain. Our experiments are conducted only on CT-RATE chest CT data, and broader validation across institutions, anatomies, and imaging modalities is left for future work. 
% In addition, the pathology classifier operates on a fixed set of 18 findings, which may limit coverage of rare or open-vocabulary abnormalities. 
The retrieval module is similarly constrained by the diversity of the training corpus.
% Our evaluation primarily relies on automated clinical metrics and LLM-based assessment rather than expert radiologist review. 
Finally, CLarGen remains dependent on the quality of current 3D medical encoders.
Potential risks include automation bias: fluent outputs may cause radiologists to under-scrutinize missed rare findings, and distribution shift, as performance may degrade on different scanner types or patient populations not represented in the training data.

\section*{Acknowledgments}
This work was supported by the Munich Center for Machine Learning (MCML), the German Research Foundation (DFG), the DAAD programme Konrad Zuse Schools of Excellence in Artificial Intelligence, sponsored by the Federal Ministry of Research, Technology and Space, and BMWi (project “NeuroTEMP”) research funding. We gratefully thank the scientific support and resources of the AI service infrastructure LRZ AI Systems provided by the Leibniz Supercomputing Centre (LRZ) of the Bavarian Academy of Sciences and Humanities (BAdW), funded by Bayerisches Staatsministerium für Wissenschaft und Kunst (StMWK).

\bibliography{paper}

\clearpage
\appendix
\section{Appendix}
\label{sec:appendix}

\renewcommand{\figurename}{Appendix Fig.}
\renewcommand{\thefigure}{A.\arabic{figure}}
\renewcommand{\theHfigure}{A.\arabic{figure}}
\renewcommand{\tablename}{Appendix Table}
\renewcommand{\thetable}{A.\arabic{table}}
\renewcommand{\theHtable}{A.\arabic{table}}
\setcounter{figure}{0}
\setcounter{table}{0}

\subsection{Pathology Labels}
\label{app:pathology_labels}

Our pathology classifier predicts 18 thoracic findings commonly observed in chest CT examinations. The predicted labels are:
\textit{medical material},
\textit{arterial wall calcification},
\textit{cardiomegaly},
\textit{pericardial effusion},
\textit{coronary artery wall calcification},
\textit{hiatal hernia},
\textit{lymphadenopathy},
\textit{emphysema},
\textit{atelectasis},
\textit{lung nodule},
\textit{lung opacity},
\textit{pulmonary fibrotic sequela},
\textit{pleural effusion},
\textit{mosaic attenuation pattern},
\textit{peribronchial thickening},
\textit{consolidation},
\textit{bronchiectasis}, and
\textit{interlobular septal thickening}.

\subsection{Normal Bias Similarity (NBS) Metric}
\label{app:nbs}
 
We introduce Normal Bias Similarity (NBS) to quantify semantic drift toward generic normal-report templates, a failure mode not reliably captured by surface-level diversity metrics such as distinct-$n$.
 
\paragraph{Formal definition.}
Let $\mathcal{N} = \{r_i\}_{i=1}^{|\mathcal{N}|}$ be a set of ground-truth reports from scans labeled as fully normal (all 18 pathology labels absent). We compute a \emph{normal reference embedding} by averaging their sentence-level embeddings:
\begin{equation}
    \bar{\mathbf{e}}_{\mathcal{N}} = \frac{1}{|\mathcal{N}|} \sum_{r_i \in \mathcal{N}} \phi(r_i),
\end{equation}
where $\phi(\cdot)$ denotes a pretrained text encoder (we use the text tower of HLIP, consistent with our retrieval setup). For each generated report $\hat{r}$, NBS is then:
\begin{equation}
    \mathrm{NBS}(\hat{r}) = \cos\!\left(\phi(\hat{r}),\, \bar{\mathbf{e}}_{\mathcal{N}}\right).
\end{equation}
We report the mean NBS over the full test set. A high mean NBS indicates that generated reports are semantically close to a prototypical normal report, regardless of the actual pathological content of the input scan---the hallmark of template collapse.
 
\paragraph{Relationship to other diversity metrics.}
Distinct-$n$ and effective vocabulary size (effV$_n$) measure lexical diversity at the corpus level; they can be high even when every report describes the same normal findings using slightly varied wording. NBS is complementary: it measures semantic proximity to a specific degenerate mode (the normal template) in embedding space, and is therefore sensitive to meaning-level collapse that surface metrics miss.

\subsection{Prompt Templates}
\label{app:prompts}

We provide the prompts used for report generation in CLarGen. The system prompt defines the clinical reporting behavior and formatting constraints, while the user prompt supplies predicted pathology labels and retrieved reference reports.

\begin{figure*}[t]
\small
\begin{mdframed}[backgroundcolor=gray!5,linewidth=0.5pt]
\textbf{System Prompt}

\begin{verbatim}
You are an AI assistant specialized in radiology, tasked with
generating clinical reports. Your writing must be indistinguishable
from that of an expert human radiologist.

Your task is to synthesize a report based on two sources of
information:
1. Expert Findings: A definitive list of pathologies found to be
   'present' or 'absent' in the scan.
2. Reference Reports: A set of reports from similar clinical cases.

Operational rules:
- Output only two sections: Findings : and Impressions :
- Do not contradict the expert labels or invent findings.
- Use the reference reports to guide vocabulary and phrasing.
- Maintain concise and objective radiology style.
- Avoid verbose or assistant-like language.

The report is evaluated for:
1. Clinical accuracy.
2. Stylistic fidelity.

Strict Output Format:
Your response must only contain the Findings : and Impressions :
sections.
\end{verbatim}
\end{mdframed}

\vspace{0.5em}

\begin{mdframed}[backgroundcolor=gray!5,linewidth=0.5pt]
\textbf{User Prompt Template}

\begin{verbatim}
Here is the information for a new CT scan.
Please write the radiology report.

Expert Findings for the Scan:
{expert_labels}

Reference Reports from Similar Cases:
{top_k_reports}

Your Generated Report:
\end{verbatim}
\end{mdframed}

\caption{Prompts used for report generation in CLarGen.}
\label{fig:prompt_templates}
\end{figure*}

\subsection{Comparison of generated reports}

We present a qualitative comparison between a ground-truth radiology report and the corresponding report generated by CLarGen for the same CT scan in Fig.~\ref{fig:reports_comparison}. The example is selected to illustrate a case with multiple co-occurring pathologies spanning several severity levels, including both common findings (cardiomegaly, vascular calcification) and rarer ones (interstitial changes, consolidation, pleural effusion).
 
Color coding highlights clinical findings that appear in both the ground truth and the CLarGen output, grouped by finding category: \hlg{green} = medical material / devices; \hlo{orange} = cardiac size; \hlp{pink} = vascular calcifications; \hly{yellow} = pleural or pericardial effusion; \hlk{blue} = lymphadenopathy; \hlm{purple} = parenchymal and interstitial findings.
 
Several observations are worth noting. First, CLarGen correctly identifies and describes all major pathological categories present in the ground truth, including the rarer interstitial and parenchymal findings (\hlm{purple}) that VLMs systematically suppress. Second, the generated report adopts realistic radiology phrasing and structure: Findings followed by a clinically consistent Impression, without having been fine-tuned on the dataset. Third, while some surface-level phrasing naturally differs from the reference (as expected from a frozen LLM guided by exemplars rather than a copying mechanism), the clinical content is preserved with high fidelity. This stands in contrast to VLM baselines, which produce reports with higher lexical overlap to common normal templates but critically miss pathological findings, as evidenced by the quantitative results in Table~\ref{tab:all_results}.

\begin{figure*}[h!]
\small
\setlength{\tabcolsep}{3pt}
\begin{tabularx}{\textwidth}{|X|X|}
\hline
\textbf{Ground Truth} & \textbf{CLarGen} \\
\hline
        
Examination secondary to breathing movements was evaluated as suboptimal.
\hlg{Tracheostomy follows}. Trachea, both main bronchi are open.
Mediastinal main vascular structures are normal.
\hlo{Heart size increased}. Pericardial effusion-thickening was not observed.
\hlp{Calcific atheroma plaques are observed in the coronary arteries and aortic arch}.
\hly{Pericardial effusion is present in minimal plastering style}.
\hlk{A few lymph nodes measuring up to 15 mm are observed in the mediastinum and hilar regions}.
Thoracic esophagus calibration was normal and no significant tumoral wall thickening was detected.
Bilateral axillary lymph node enlarged in pathological dimensions was not detected.
\hly{A small amount of effusion is observed in both hemithorax, more prominent on the right}.
When examined in the lung parenchyma window;
\hlm{Clarification of interstitial signs in both lungs},
\hlm{budding tree images} being more prominent in the lower lobe of the right lung,
and \hlm{an increase in density consistent with the consolidation of 21 mm in the basal segment of the lower lobe of the left lung} are observed.
        
& 
        
\hlo{CTO slightly increased in favor of the heart}. There is local thickening of the pericardium. [...]
Calibration of the aortic arch and other major vascular structures is natural.
\hlp{Calcific atheroma plaques are observed in the ascending and descending aorta in the aortic arch, and in the coronary arteries}.
\hlk{In almost all lymph node groups in the mediastinum, lymph nodes are present in the right lower paratracheal area, with a central slightly hypodense appearance}.
Densities that partially level in the esophagus are observed.
\hlk{There are lymph nodes at the hilar level}.
When examined in the lung parenchyma window; Both hemithorax are symmetrical.
There is a catheter view inside the trachea.
\hlm{Consolidative areas with atelectasis-air bronchograms are observed in both lungs}, extending from the basal to the apex,
in the vicinity of the \hly{pleural effusion is the most prominent on the right and also on the left}.
The described findings are also observed in the upper lobe posterior segments and lower lobe superior segments in both lungs.
It is recommended to be evaluated for aspiration pneumonia.
\hlm{There are thickenings in the common interlobular septa}, more prominent in the upper lobes.
        
\\
\hline
\end{tabularx}
\caption{Qualitative comparison of a ground-truth report and the CLarGen-generated report for the same CT scan. Color coding highlights findings by category: \hlg{green} = devices/medical material; \hlo{orange} = cardiac; \hlp{pink} = vascular calcifications; \hly{yellow} = effusions; \hlk{blue} = lymphadenopathy; \hlm{purple} = parenchymal/interstitial findings. CLarGen recovers all major pathological categories present in the ground truth, including rare interstitial and consolidative changes, using clinically appropriate terminology and report structure.}
\label{fig:reports_comparison}
\end{figure*}

\subsection{Additional Quantitative Analysis on Template Collapse}
\label{app:template_collapse_evidence}

\subsubsection{Dominant Template Analysis}
\label{app:dominant_templates}

Table~\ref{tab:dominant_templates} directly visualizes template collapse by examining the most frequently generated report for each method on the 1,564-case common subset. In the ground truth, the single most common report appears only 13 times (0.8\%), reflecting the natural variation expected across patient-specific findings. VLM baselines deviate strikingly from this pattern. CT-CHAT emits its most frequent template 452 times, covering 28.9\% of the entire subset, and SAMF 325 times (20.8\%). Crucially, both templates describe an entirely normal chest: no pathological findings, no abnormalities. This means that for nearly one in three cases, CT-CHAT produces \emph{the exact same normal report} regardless of what pathology is present in the scan. Combined with Surv$_\text{rare}$ scores of 0.035 and 0.000, respectively (Table~\ref{tab:all_results}), this confirms that template collapse is not a statistical artifact of aggregate metrics but a concrete, case-level failure.

CLarGen's most frequent output covers only 28 cases (1.8\%), on par with the ground-truth reference rate. Notably, even this most common CLarGen template is more conservative than those of the VLM baselines: it explicitly acknowledges diagnostic uncertainty due to the non-contrast examination rather than asserting normality, reflecting the clinical hedging characteristic of real radiology practice.

\begin{table*}[t]
\centering
\caption{Dominant exact templates on the 1,564-case common subset. Counts show how often the most frequent normalized report appears. CT-CHAT and SAMF repeatedly emit long normal-template reports, whereas the most frequent CLarGen report covers only 1.8\% of the subset.}
\label{tab:dominant_templates}
\begin{tabular}{p{2.2cm} cc p{9.5cm}}
\toprule
\textbf{Model} & \textbf{Count} & \textbf{Share} & \textbf{Most frequent normalized report template} \\
\midrule

Ground truth & 13 & 0.8\% &
Findings: Trachea, both main bronchi are open. Mediastinal main vascular structures, heart contour, size are normal. Thoracic aorta diameter is normal. Pericardial effusion-thickening was not observed. Thoracic esophageal calibration was normal and no significant tumoral wall thickening was detected. [...] \\

CT-CHAT & 452 & 28.9\% &
Findings: Trachea, both main bronchi are open. Mediastinal main vascular structures, heart contour, size are normal. Thoracic aorta diameter is normal. Pericardial effusion-thickening was not observed. Thoracic esophagus calibration was normal and no significant tumoral wall thickening was detected. [...] \\

SAMF & 325 & 20.8\% &
Trachea and both main bronchi were in the midline and no obstructive pathology was detected in the lumen. The mediastinum could not be evaluated optimally in the non-contrast examination. As far as can be observed: Mediastinal main vascular structures, heart contour and size are normal. [...] \\

CLarGen & 28 & 1.8\% &
Findings: Mediastinal structures were evaluated as suboptimal because the examination was unenhanced. As far as can be seen; trachea and main bronchi are open. No pathological LAP was detected in the mediastinum. The heart and mediastinal vascular structures are normal. [...] \\

\bottomrule
\end{tabular}
\end{table*}

\begin{table}[h!]
\centering
\caption{Model-level association between surface-overlap metrics and clinical F1 on the common subset, excluding ground truth. Negative correlations indicate that higher lexical overlap does not imply better clinical fidelity.}
\label{tab:metric_f1_correlation}
\begin{tabular}{lcc}
\toprule
Metric vs.\ F1 & Pearson & Spearman \\
\midrule
ROUGE-L   & -0.324 & -0.400 \\
BLEU-mean & -0.850 & -0.600 \\
METEOR    & -0.914 & -0.700 \\
BLEU-4    & -0.782 & -0.900 \\
\bottomrule
\end{tabular}
\end{table}

% Added concise report-level template diagnostics to quantify the dominant-template pattern.
\begin{table}[h!]
\centering
\caption{Report-level template concentration on the common subset. Unique is the fraction of distinct normalized reports. Top-1 and Top-5 measure coverage by the most frequent templates, and T80 counts templates needed to cover 80\% of outputs.}
\setlength{\tabcolsep}{2.5pt}
\label{tab:tc_diagnostics}
\begin{tabular}{lcccc}
\toprule
Model & Unique $\uparrow$ & Top-1 $\downarrow$ & Top-5 $\downarrow$ & T80 $\uparrow$ \\
\midrule
Ground truth & 0.969 & 0.008 & 0.026 & 1204 \\
CT-CHAT      & 0.201 & 0.289 & 0.478 & 59 \\
HULU-Med     & 0.992 & 0.007 & 0.012 & 1239 \\
Reg2Rg       & 0.953 & 0.003 & 0.012 & 1179 \\
SAMF         & 0.012 & 0.208 & 0.734 & 6 \\
CLarGen      & 0.698 & 0.018 & 0.056 & 780 \\
\bottomrule
\end{tabular}
\end{table}

% Added short sentence-level analysis without adding a full extra table.
% Sentence-level clustering gives the same qualitative diagnosis when exact report duplication is too strict.
Sentence-level clustering gives the same qualitative diagnosis when exact matching underestimates reuse.
RadBERT sentence clusters show that CT-CHAT and SAMF require only 39 and 34 clusters to cover 80\% of generated sentences, compared with 62 for ground truth and 54 for CLarGen.
This indicates that template collapse also appears as repeated sentence-level clinical fragments, not only as exact full-report copies.

\subsubsection{NLP and clinical score correlation}
% Table~\ref{tab:metric_f1_correlation} reports the Pearson and Spearman correlations between standard NLP metrics and macro-F1 across all five models on the common subset. 
% Across the five evaluated models, all correlations are negative, consistent with the template collapse hypothesis: models achieving higher lexical overlap tend to do so by conforming to frequent normal-report patterns rather than capturing pathological content. 
% The consistency across five independent metrics (Pearson range $-0.324$ to $-0.938$) corroborates the qualitative pattern observed in Table~\ref{tab:all_results}.
Table~\ref{tab:metric_f1_correlation} reports the Pearson and Spearman correlations between standard surface-overlap metrics and macro-F1 on the common subset.
All four correlations are negative, showing that higher lexical overlap does not imply better clinical fidelity in this setting.
This supports the template-collapse diagnosis in Table~\ref{tab:all_results}: fluent reports can match common reference phrasing while omitting patient-specific findings.

\begin{table*}[b]
\centering
\caption{Rare-finding survival on the 1,564-case common subset. Survival is the fraction of ground-truth-positive (GT+) cases for which the generated report still contains the pathology according to RadBERT extraction. We present six rarest pathologies in the evaluation set.}
\label{tab:rare_finding_survival}
\begin{tabular}{lcccccc}
\toprule
Pathology & GT+ & CT-CHAT & HULU-Med & Reg2Rg & SAMF & CLarGen \\
\midrule
Pericardial effusion             & 112 & 0.009 & 0.054 & 0.000 & 0.000 & 0.714 \\
Interlobular septal thickening   & 120 & 0.050 & 0.083 & 0.000 & 0.000 & 0.675 \\
Mosaic attenuation pattern       & 126 & 0.040 & 0.079 & 0.040 & 0.000 & 0.595 \\
Medical material                 & 157 & 0.006 & 0.076 & 0.191 & 0.000 & 0.363 \\
Cardiomegaly                     & 164 & 0.043 & 0.006 & 0.024 & 0.000 & 0.762 \\
Peribronchial thickening         & 171 & 0.012 & 0.099 & 0.012 & 0.000 & 0.673 \\
\bottomrule
\end{tabular}
\end{table*}

\subsubsection{Rare-finding survival}

Table~\ref{tab:rare_finding_survival} reports per-finding survival rates for the six rarest pathologies in the evaluation set, alongside the number of ground-truth positive cases (GT+) for each. The GT+ column is critical for contextualizing the severity of suppression: pericardial effusion, for instance, is present in 112 scans, yet SAMF and Reg2Rg achieve survival rates of exactly 0.000, meaning neither model mentions pericardial effusion in a single case across the entire test set. CT-CHAT reaches 0.009, corresponding to roughly one correct mention per 111 positive scans. 
CLarGen recovers a substantial fraction of these findings across all six pathologies (0.363-0.762), confirming that the decoupled detection objective directly addresses the suppression mechanism. The one relatively lower CLarGen survival, Medical material, at 0.363, is consistent with the fact that medical devices are among the hardest findings to detect from volumetric features alone, as their appearance is highly variable and scanner-dependent.

\subsection{Image-Input VLM Ablation: Template Collapse Under Visual Conditioning}
\label{app:sec:image-input-ablation}

A natural question is whether directly conditioning the report generator on HLIP image embeddings, bypassing the discrete LQT bottleneck, would improve clinical completeness. We test two such variants: \textbf{HLIP$\rightarrow$Qwen} and \textbf{HLIP$\rightarrow$Llama}, which feed frozen HLIP representations as visual tokens to Qwen2.5-VL~\cite{bai2025qwen3} and Llama-3.2-Vision~\cite{touvron2023llama}, respectively, and fine-tune them to decode reports autoregressively. All models are evaluated on the same CT-RATE validation set; labels are extracted from generated reports with a RadBERT-based extractor and compared against the CT-RATE ground truth.

\begin{table*}[t]
\centering
% \small
\setlength{\tabcolsep}{6pt}
\caption{Clinical and template-collapse metrics on the CT-RATE validation set. \textit{Drop rate}: fraction of GT-positive labels absent from the generated report. \textit{Gen.\ Pos./vol}: mean positive findings per report. \textit{Unique Rpt.\ Rate}: fraction of reports that are textually distinct. \textit{Top-1 Share}: fraction of volumes assigned the single most frequent report. \textit{T80}: number of unique reports needed to cover 80\% of the corpus.}
\label{tab:vlm_ablation}
\begin{tabular}{lcccccccc}
\toprule
\textbf{Model} & \textbf{F1}$\uparrow$ & \textbf{Rec.}$\uparrow$ & \textbf{Prec.}$\uparrow$ & \textbf{Drop}$\downarrow$ & \textbf{Gen.\ P/v} & \textbf{Uniq.\ \%}$\uparrow$ & \textbf{Top-1}$\downarrow$ & \textbf{T80}$\uparrow$ \\
\midrule
LQT (original) & 0.519 & 0.759 & 0.414 & 0.255 & 6.06 & --- & --- & --- \\
CLarGen        & 0.497 & 0.663 & 0.416 & 0.344 & 5.25 & 0.691 & 0.017 & 789 \\
HLIP$\rightarrow$Qwen  & 0.240 & 0.189 & 0.463 & 0.755 & 1.51 & 0.373 & 0.252 & 278 \\
HLIP$\rightarrow$Llama & 0.316 & 0.263 & 0.499 & 0.685 & 1.97 & 0.530 & 0.176 & 530 \\
\bottomrule
\end{tabular}
\end{table*}

\begin{table*}[h!]
\centering
\caption{Flow decomposition relative to the LQT detector. TP keep/drop measures whether detector true positives survive in the generated report. FP propagate/correct measures whether detector false positives are repeated or removed.}
\label{tab:vlm_flow}
\begin{tabular}{lcccc}
\toprule
Model & TP keep $\uparrow$ & TP drop $\downarrow$ & FP prop. $\downarrow$ & FP corr. $\uparrow$ \\
\midrule
CLarGen & 0.879 & 0.121 & 0.844 & 0.156 \\
HLIP$\rightarrow$Qwen & 0.307 & 0.693 & 0.148 & 0.852 \\
HLIP$\rightarrow$Llama & 0.404 & 0.596 & 0.213 & 0.787 \\
\bottomrule
\end{tabular}
\end{table*}

\begin{figure}[h!]
\centering
\includegraphics[width=0.95\linewidth]{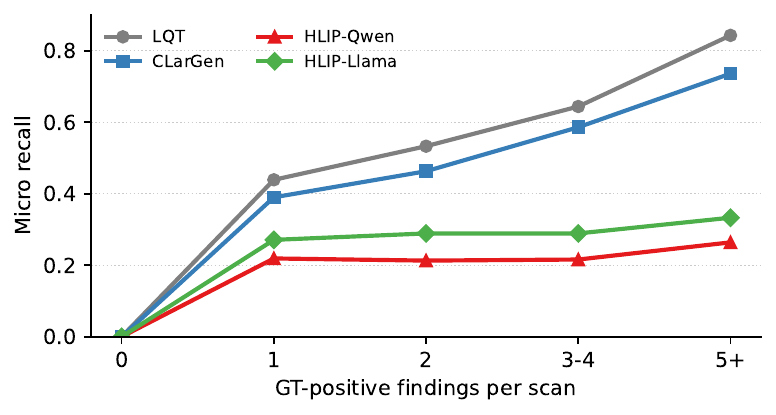}
\caption{Disease-burden sensitivity for image-input VLMs. As the number of ground-truth-positive findings increases, image-input VLM recall remains low, while CLarGen preserves substantially more positive findings.}
\label{fig:vlm_burden_recall}
\end{figure}

% The image-input VLMs are severely omission-prone: HLIP$\rightarrow$Qwen and HLIP$\rightarrow$Llama generate only 1.51 and 1.97 positive findings per report against 3.43 GT positives per volume, yielding macro F1 of 0.240 and 0.316 compared with 0.497 for CLarGen. The diversity metrics reveal a complementary failure mode we term \textit{template collapse}: over 25\% of HLIP$\rightarrow$Qwen reports are identical copies of a single template, and only 278 unique reports are needed to cover 80\% of the corpus (vs.\ 789 for CLarGen), indicating that the autoregressive decoder converges to a small set of conservative, finding-sparse outputs. Notably, the higher precision of the image-input VLMs (0.463/0.499 vs.\ 0.416) is not a sign of selective accuracy but rather a byproduct of blanket suppression: by generating very few positive findings, the models trivially avoid false positives at the cost of missing the majority of true pathology.

% These results demonstrate that template collapse is not caused by the absence of visual input at generation time. Even when provided with the same HLIP features that drive CLarGen's upstream classifier, autoregressive report decoders suppress weak or ambiguous clinical evidence in favor of fluent, conservative text. CLarGen's retrieve-and-prompt design sidesteps this failure mode by externalizing clinical decisions to the discrete LQT stage and grounding generation in retrieved exemplar reports.

The image-input VLMs are severely omission-prone: HLIP$\rightarrow$Qwen and HLIP$\rightarrow$Llama generate only 1.51 and 1.97 positive findings per report against 3.43 GT positives per volume, yielding macro-F1 of 0.240 and 0.316 compared with 0.497 for CLarGen.
Their higher precision is a byproduct of suppression rather than selective clinical accuracy: Table~\ref{tab:vlm_flow} shows that they drop 59.6--69.3\% of LQT true positives, while CLarGen preserves 87.9\%.
The diversity metrics reveal the accompanying template-collapse pattern: over 25\% of HLIP$\rightarrow$Qwen reports are identical copies of a single template, and only 278 unique reports are needed to cover 80\% of the corpus, compared with 789 for CLarGen.
Figure~\ref{fig:vlm_burden_recall} shows the same failure under increasing disease burden, where image-input VLM recall remains low even as the number of ground-truth findings grows.
Visual input alone therefore does not prevent template collapse; autoregressive report decoders still suppress weak clinical evidence unless the pipeline makes clinical content explicit before generation.

\subsection{Computational Cost}
\label{sec:computational_cost}
CLarGen is designed to be computationally lightweight at both training and inference time. The only trainable components are the LQT classifier and the retrieval projector, which comprise approximately 10K parameters; the HLIP encoder and MedGemma-27B are kept entirely frozen throughout training. This design makes CLarGen practical to train without requiring large-scale GPU resources or fine-tuning a generative model.
At inference time, CLarGen operates on precomputed HLIP image embeddings and executes the full pipeline: LQT classification, evidence retrieval, and MedGemma-27B report generation, in a mean of 1.5 seconds per report on a single NVIDIA H100 GPU, with a peak memory footprint of approximately 90GB, dominated by the frozen MedGemma-27B backbone. These demonstrate that CLarGen achieves strong clinical report quality without sacrificing deployment efficiency. The approximately 90GB peak memory footprint corresponds to full-precision MedGemma-27B inference and reflects our experimental configuration rather than a deployment requirement of CLarGen. The same model requires approximately 15~GB in 4-bit precision, and smaller LLM backbones can be substituted without retraining the preceding stages.

\end{document}